\PassOptionsToPackage{numbers,compress}{natbib}
\documentclass{article}

\usepackage[preprint,sglblindworkshop]{neurips_2025}

\workshoptitle{Learning from Time Series for Health (TS4H)}

\makeatletter
\renewcommand{\@noticestring}{%
  Workshop on Learning from Time Series for Health, \@neuripsordinal\ Conference on Neural Information Processing Systems (NeurIPS \@neuripsyear).%
}
\makeatother

\usepackage[utf8]{inputenc}
\usepackage[T1]{fontenc}
\usepackage{hyperref}
\usepackage{url}
\usepackage{booktabs}
\usepackage{amsmath, amssymb}
\usepackage{nicefrac}
\usepackage{microtype}
\usepackage{xcolor}
\usepackage{graphicx}
\usepackage{multirow}
\usepackage{capt-of}
\usepackage{float}
\usepackage{placeins}
\usepackage{dblfloatfix}

\usepackage{caption}
\title{AttentiveGRUAE: An Attention-Based GRU Autoencoder for Temporal Clustering
and Behavioral Characterization of Depression from Wearable Data}

\author{
  Nidhi Soley\thanks{Corresponding author: \texttt{nsoley1@jhu.edu}} \\
  Department of Biomedical Engineering, \\
  Institute for Computational Medicine \\
  Johns Hopkins University \\
  Baltimore, MD \\
  \And
  Vishal M.\ Patel \\
  Department of Electrical and Computer Engineering \\
  Johns Hopkins University \\
  Baltimore, MD
  \And
  Casey O.\ Taylor \\
  Departments of Medicine and Biomedical Engineering, \\
  Institute for Computational Medicine \\
  Johns Hopkins University \\
  Baltimore, MD
}
\begin{document}
\maketitle
\raggedbottom

\begin{abstract}
In this study, we present AttentiveGRUAE, a novel attention-based gated recurrent unit (GRU) autoencoder designed for temporal clustering and prediction of outcome from longitudinal wearable data. Our model jointly optimizes three objectives: (1) learning a compact latent representation of daily behavioral features via sequence reconstruction, (2) predicting end-of-period depression rate through a binary classification head, and (3) identifying behavioral subtypes through the Gaussian Mixture Model (GMM) based soft clustering of learned embeddings. We evaluate AttentiveGRUAE on longitudinal sleep data from 372 participants (GLOBEM 2018–2019), and it demonstrates superior performance over baseline clustering, domain-aligned self-supervised, and ablated models in both clustering quality (silhouette score = 0.70 vs (0.32–0.70)) and depression classification (AUC = 0.74 vs (0.50–0.67)). Additionally, external validation on cross-year cohorts from 332 participants (GLOBEM 2020–2021) confirms cluster reproducibility (silhouette score = 0.63, AUC = 0.61) and stability. We further perform subtype analysis and visualize temporal attention, which highlights sleep-related differences between clusters and identifies salient time windows that align with changes in sleep regularity, yielding clinically interpretable explanations of risk. 
\end{abstract}

\section{Introduction}
Wearable sensing provides longitudinal, real-world behavioral signals relevant to depression risk. Yet, many modeling approaches assume dense, high-frequency multimodal data or they function as black-box classifiers with limited clinical interpretability \citep{Fried2015Heterogeneity, xu2023globem, Mobile}. In practice, low-frequency daily summaries are widely available and clinically salient: short or irregular sleep is a modifiable risk factor repeatedly linked to depression \citep{Murphy2015SleepDepression, Wang2024NHANES, Lim2022SleepEfficiency, Li2016InsomniaMeta}. Three known challenges for time-series health models are: (i) outcome-agnostic clustering that may not align with clinical screening endpoints \citep{Wang2024ClinicalTemporalClustering}; (ii) poor temporal interpretability about when risk meaningfully diverges within a trajectory; and (iii) lack of reproducibility.

To address these gaps, we propose AttentiveGRUAE, a framework for interpretable and reproducible temporal clustering of time series wearable data. AttentiveGRUAE is trained end-to-end to jointly optimize (1) a sequence reconstruction loss to capture temporal dynamics, and (2) a binary outcome loss to ensure that representations are clinically informative. We evaluate this model on the public GLOBEM dataset \citep{xu2023globem}, which comprises four cohorts of wearable data collected across different institutions and years. Our model is benchmarked against several baselines, including various ablations, traditional time series clustering models, and domain-aligned time-series baselines. We further examine cluster reproducibility and stability across multi-year cohorts and analyze behavioral profiles to provide clinically meaningful interpretations.

\textbf{Contributions.} (1) An interpretable, outcome-aware temporal clustering framework that couples attention-guided sequence encoding with soft subtyping for day-level insight; (2) comprehensive benchmarking against both domain-aligned baselines and classical ML methods, with ablative analyses of attention and joint training; and (3) reproducibility and stability demonstrated via cross-cohort (pre/post-COVID) validation and resampling-based Adjusted Rand Index (ARI), yielding consistent subtype structure.

\section{Related Work}
Passive sensing with wearables and smartphones has linked sleep regularity, mobility, and phone use to depressive symptoms, motivating machine learning approaches for screening from behavioral time series data \citep{saeb2015mobile,ben-zeev2015next,wang2018tracking,xu2019leveraging,xu2021leveraging,chikersal2021detecting}. While these studies established feasibility, many treat depression as a cross-sectional classification problem and do not discover clinically meaningful subtypes \citep{farhan2016behavior,xu2023globem}. In parallel, deep sequence models (e.g., RNN/CNN/GNN variants) have advanced representation learning in health data \citep{rajkomar2018scalable,lee2020temporal,catling2020temporal,han2023multi,machado2024identifying}, and unsupervised deep temporal clustering has been explored for latent structure discovery \citep{madiraju2018deep,zhong2021deep,ezugwu2022survey}; however, outcome-agnostic clustering may misalign with screening endpoints and often lacks temporal interpretability. In the work by \citet{Wang2024ClinicalTemporalClustering}, outcome-guided subtyping (OG-DTC) addresses alignment but relies on hard assignments with limited interpretability. Attention-augmented GRU autoencoders have been used in high-frequency physiological signals (e.g., ECG) for supervised detection \citep{Roy2024AttentivECGRU}, a setting that differs from daily wearable summaries in sampling rate and task design. Self-supervised methods such as time-series representation learning via temporal and contextual contrasting (TS-TCC) \citep{Eldele2021TSTCC} and broader time-series foundation models \citep{Liang2024TSSurvey} primarily target dense, high-frequency multivariate data. 

In contrast to outcome-agnostic clustering and high-frequency waveform settings, the focus here is on low-frequency wearable sleep summaries, with comparisons to domain-aligned self-supervised baselines and evaluation under temporal distribution shifts for reproducibility.

\section{Methods}
\begin{figure}[H]
\centering
\includegraphics[width=0.70\linewidth]{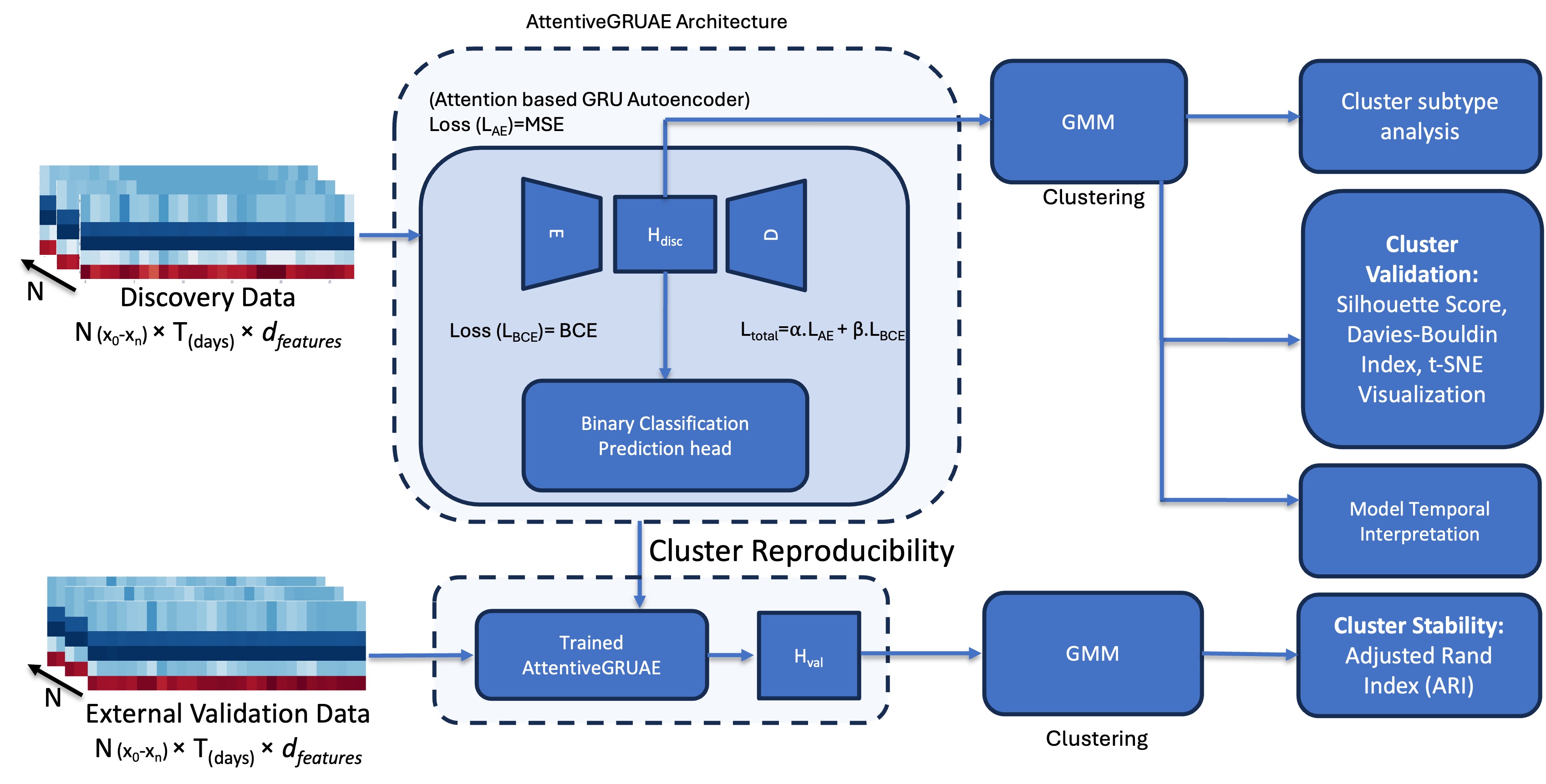}
\caption{\small Outcome Guided-GRU encoder–decoder with attention; embeddings clustered by GMM (BIC for $K$). We report AUC/MSE, Silhouette/DBI, and assess cluster reproducibility via ARI on external validation data.}
\label{fig:overview}
\end{figure}
\noindent\textbf{Overview.}
We propose \emph{AttentiveGRUAE}, an interpretable framework for temporal clustering of behavioral time series. Given $X=\{x_i\}_{i=1}^N$ with $x_i\in\mathbb{R}^{T\times d}$ (a $T$-day sequence of $d$ wearable-derived features) and a binary outcome $y_i\in\{0,1\}$, the model jointly learns to (i) reconstruct input sequences and (ii) predict the outcome, producing participant-level embeddings that are subsequently clustered into behavioral subtypes (Fig.~\ref{fig:overview}).

\noindent\textbf{Architecture and objective.}
AttentiveGRUAE uses a GRU encoder with multi-head attention over encoder states, a GRU decoder for sequence reconstruction, and a small feed-forward head for binary outcome prediction. The encoder produces a participant-level embedding that is pooled across time and shared by both heads: the decoder reconstructs the sequence, and the prediction head estimates the end-of-window outcome label. Training optimizes a weighted sum of a reconstruction loss (mean-squared error over all time steps and features) and a prediction loss (binary cross-entropy of the classification head). The architectural details and mathematical definitions are provided in the Supplementary Material S1.

\noindent\textbf{Training and clustering.}
We train end-to-end with Adam optimizer \citep{kingma2014adam}, gradient clipping, dropout, and $L_2$ regularization; early stopping monitors validation AUC, and a ReduceLROnPlateau scheduler lowers the learning rate on plateaus (max 50 epochs, batch size 64). To mitigate occasional gradient conflict between tasks, we apply gradient surgery during joint training. After training, we extract latent embeddings $h_i$ per participant and fit a GMM with $K$ selected by BIC (Supplementary Fig.~4), yielding soft subtype assignments (supplement S2). We report MSE/AUC to quantify representation quality, Silhouette and Davies–Bouldin for cluster structure, and t-SNE qualitatively. For external validation, we freeze the encoder, compute $H_{\text{val}}$ on a held-out cohort, and assess stability via ARI.

\noindent\textbf{Experiment dataset and features.}
We use the public GLOBEM dataset \citep{xu2023globem} (705 participants) with a pre-COVID discovery split (DS1+DS2; 2018–2019; $n{=}373$) and a post-COVID validation split (DS3+DS4; 2020–2021; $n{=}332$). Inclusion requires $\ge$7 days of wearable data in a trailing 28-day window and a depression label at the end of the 28-day window. Outcomes follow GLOBEM’s clinical cut points (PHQ-4\,$>$\,2 or BDI-II\,$>$\,13). Inputs are six sleep summaries (duration, efficiency, latency, first bedtime/waketime, plus daily average duration), extracted daily, imputed using forward/backward fill, windowed to 28 days, and standardized using training-split statistics.

\noindent\textbf{Baselines and ablations.}
Most recent foundation models target dense, high-frequency modalities; we therefore report (i) conventional baselines (PCA\,$+$\,GMM, $k$-means, time-series $k$-means) and (ii) a domain-aligned self-supervised baseline: TS-TCC \citep{Eldele2021TSTCC} re-implemented from scratch on the same sleep-only discovery split, no pretraining, with the identical downstream pipeline (frozen encoder, GMM, same prediction head). We also ablate attention and the prediction head to isolate contributions of attention and multi-task learning. OG-DTC \citep{Wang2024ClinicalTemporalClustering} is relevant but lacks public code; our AE-only ablation approximates its latent-space objective.

\section{Results}

\subsection{Performance on discovery data and ablation}
\noindent
\begin{minipage}[t]{0.48\linewidth}
Table~\ref{tab:disc} shows that the full model achieves the best overall trade-off (MSE=$0.47$, AUC=$0.74$, Silhouette=$0.70$, DBI=$0.33$), and joint training with attention is critical: removing attention reduces AUC to $0.66$; removing outcome guidance (AE-only) weakens cluster structure (Silhouette $0.52$, DBI $1.70$). TS-TCC, a strong self-supervised baseline on the same inputs, attains AUC $0.67$ and Silhouette $0.45$, but trails AttentiveGRUAE, suggesting that attention and outcome-informed objectives produce more clusterable, task-relevant embeddings for low-frequency sleep time series. Classical clustering methods underperform on both separability and prediction.
\end{minipage}\hfill
\begin{minipage}[t]{0.50\linewidth}
\centering\footnotesize
\setlength{\tabcolsep}{1.5pt}\renewcommand{\arraystretch}{1.1}
\captionof{table}{\small Discovery (DS1+DS2). MSE only for AE-based models (— = n/a). Best in \textbf{bold}.}
\label{tab:disc}
\begin{tabular}{lcccc}
\toprule
\textbf{Model} & \textbf{MSE} $\downarrow$ & \textbf{AUC} $\uparrow$ & \textbf{Sil.} $\uparrow$ & \textbf{DBI} $\downarrow$ \\
\midrule
\textbf{AttentiveGRUAE} & \textbf{0.47} & \textbf{0.74} & \textbf{0.70} & \textbf{0.33} \\
No Attention                    & 0.57          & 0.66          & \textbf{0.70} & 0.77 \\
AE-only                         & \textbf{0.47} & —             & 0.52          & 1.70 \\
Sequential Training             & 0.51          & 0.57          & 0.58          & 1.60 \\
\midrule
TS-TCC (self-sup, frozen)       & —             & 0.67          & 0.45          & — \\
PCA + GMM                       & —             & 0.50          & 0.32          & 6.39 \\
$k$-means                       & —             & 0.51          & 0.47          & 1.89 \\
Time Series $k$-means           & —             & 0.50          & 0.58          & 0.65 \\
\addlinespace
\textit{GLOBEM bench.} \citep{xu2023globem} & — & 0.51 & — & — \\
\bottomrule
\end{tabular}
\end{minipage}

\subsection{Reproducibility Across Cohorts and Cluster Stability}
\noindent
Transferring the frozen encoder to DS3+DS4 yields modest drops (Table~\ref{tab:ood}): AttentiveGRUAE retains AUC $0.61$ and Silhouette $0.63$, indicating that embeddings learned on pre-COVID data remain clusterable and predictive post-COVID. The No-Attention variant degrades substantially (AUC $0.32$, Silhouette $0.42$), underscoring attention’s role in robust temporal representation. On DS3+DS4, leave-out resampling (200 trials) yields mean ARI $0.89$ (Supplementary Fig.~5), indicating high robustness to cohort perturbations and preservation of subtype structure.
\begin{table}[t]
\centering
\footnotesize
\setlength{\tabcolsep}{30pt}\renewcommand{\arraystretch}{1.1}
\caption{Generalizability: discovery (DS1+DS2) vs.\ validation (DS3+DS4).}
\label{tab:ood}
\resizebox{\linewidth}{!}{%
\begin{tabular}{lcccc}
\toprule
\multirow{2}{*}{\textbf{Model}} & \multicolumn{2}{c}{\textbf{AUC} $\uparrow$} & \multicolumn{2}{c}{\textbf{Silhouette} $\uparrow$} \\
\cmidrule(lr){2-3}\cmidrule(lr){4-5}
 & DS1+DS2 & DS3+DS4 & DS1+DS2 & DS3+DS4 \\
\midrule
\textbf{AttentiveGRUAE (full)} & \textbf{0.74} & \textbf{0.61} & \textbf{0.70} & \textbf{0.63} \\
No Attention                    & 0.66          & 0.32          & \textbf{0.70} & 0.42 \\
AE-only                         & —             & —             & 0.52          & 0.32 \\
Sequential Training             & 0.57          & 0.54          & 0.58          & 0.28 \\
\bottomrule
\end{tabular}}
\end{table}

\subsection{Cluster subtype analysis and temporal interpretation}
\noindent
\begin{minipage}[t]{0.49\linewidth}
\centering
\includegraphics[width=\linewidth]{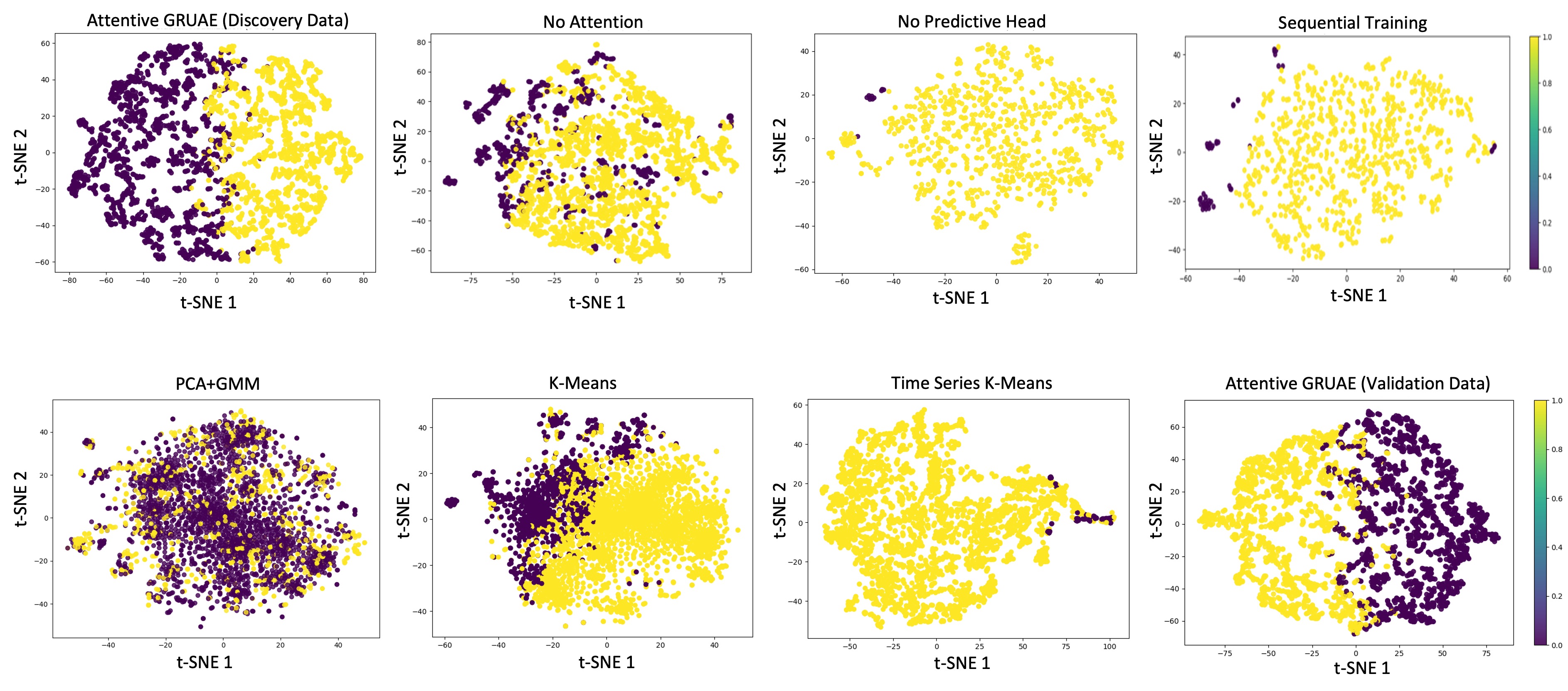}
\captionof{figure}{\small t-SNE of all the model variants and baselines (Cluster~1 = yellow, Cluster~0 = purple).}
\label{fig:clusters}
\end{minipage}\hfill
\begin{minipage}[t]{0.49\linewidth}
\centering
\includegraphics[width=\linewidth]{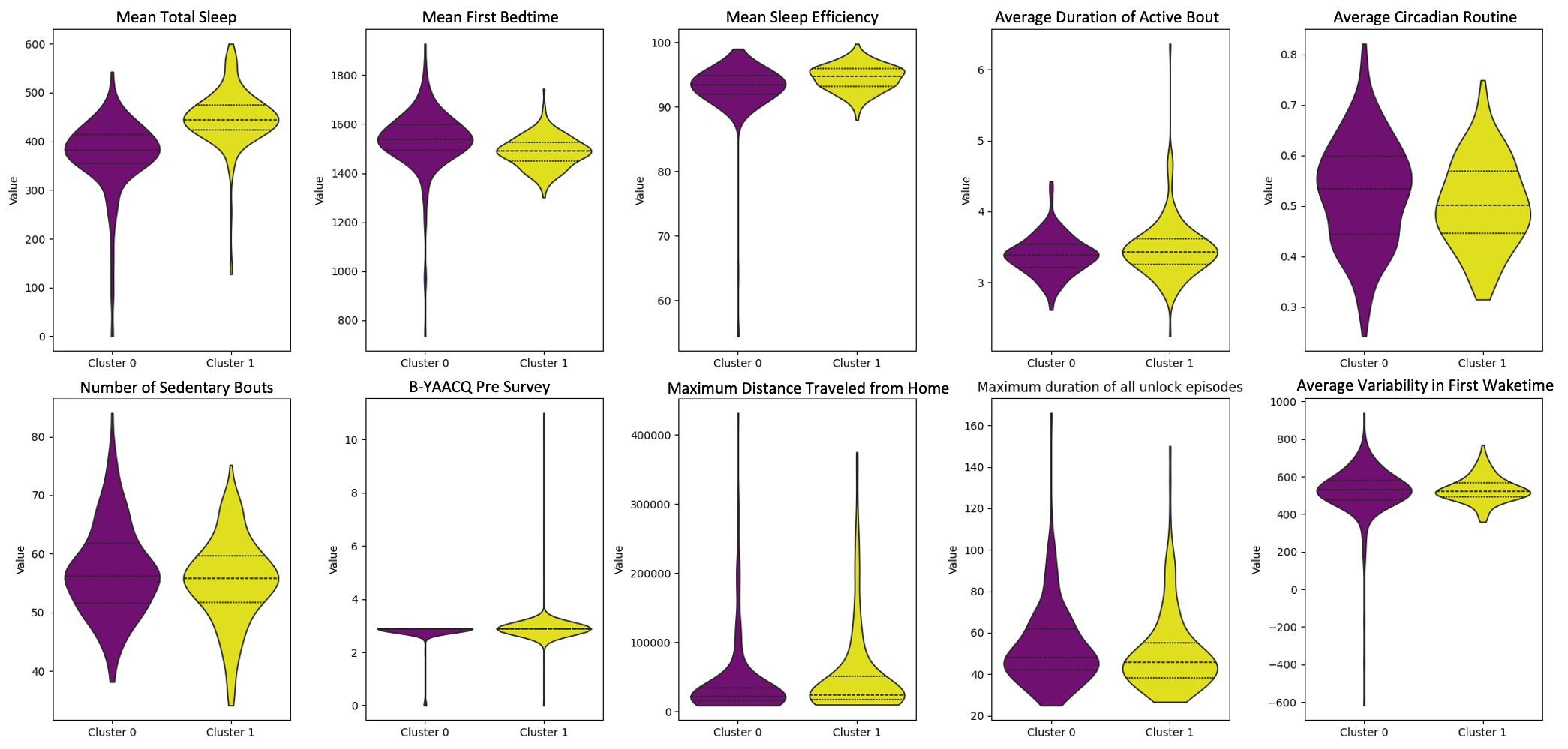}
\captionof{figure}{\small Violin plots for key differentiators.}
\label{fig:violin}\label{fig:violinplots}
\end{minipage}

\noindent AttentiveGRUAE yields clearly separated clusters on both discovery and validation cohorts (Fig.~\ref{fig:clusters}), indicating robust representations that generalize across cohorts. In contrast, a comparative t-SNE grid for ablations shows weaker and ambiguous structure, highlighting the benefit of outcome-informed attention. To quantify differences between subtypes, we ran per-feature Mann–Whitney U tests with Benjamini–Hochberg FDR correction and reported effect sizes via Cohen’s $d$ (Supplementary Figure 6, Table 3, 4). The violin plots (Fig.~\ref{fig:violin}) highlight that the dominant sleep-domain signals, Cluster~1, exhibit longer sleep durations, earlier bedtimes, and higher efficiency. 
Furthermore, the interpretation of attention weights reveals that attention peaks align with salient trajectory changes: earlier in Cluster 0 (around day 7), near sharp declines in sleep duration, and slightly later in Cluster 1 (around day 9), as sleep stabilizes; see Supplementary S3 (Figs. 7–8).

\section{Discussion}
AttentiveGRUAE couples outcome-aware learning for time series data with soft latent clustering to yield reproducible, interpretable subtypes from low-frequency wearable sleep data. On discovery, it gives the best trade-off across reconstruction, prediction, and clustering metrics. On validation data, the performance drop may reflect behavioral changes due to the pandemic. Although clustering remained stable with perturbation (ARI = 0.89), suggesting the learned subtypes are reproducible even across distribution shifts. Ablations show both components matter: removing attention sharply degrades generalization, and decoupling reconstruction/prediction weakens cluster structure. This shows that outcome guidance and temporal attention are complementary for learning behaviorally salient representations. Further, the self-supervised baseline (TS-TCC) lags in performance, indicating that attention with task supervision is better matched to low-frequency signals. Cluster subtype analysis revealed that cluster 0 comprised individuals with shorter total sleep duration, greater variability in bedtime and circadian routine, and higher depression rates. In contrast, cluster 1 includes individuals with longer, more stable sleep patterns and a lower depression rate. These findings align with well-established links between sleep and depression. Reduced sleep duration, sleep inefficiency, and irregular circadian patterns are known risk factors and correlates of depressive episodes \citep{anderson2013sleep,nutt2008sleep,franzen2008sleep,mayers2006relationship,germain2008circadian}.

Clinically, these results support \emph{risk stratification} rather than diagnosis: passively measured sleep patterns can separate behaviorally coherent subtypes that align with depression screening endpoints and remain reproducible under temporal shift. Key limitations of our study are that the evaluation is confined to a single public cohort, and sleep-only input, chosen for interpretability, may underuse available signals. Attention weights are not causal explanations; they localize influential windows but do not establish causal treatment relations. Future work will extend to multi-modal inputs and further probe causal and counterfactual relations.

\section*{Data \& Code Availability}

\textbf{Data.} All experiments use the publicly hosted \emph{GLOBEM} datasets (DS1–DS4) on PhysioNet \citep{xu2023globem-physionet}; the dataset was originally introduced in \citep{xu2022globem-neurips}. 
\textbf{Code.} Code, configs, and scripts are available in the AttentiveGRUAE repository \citep{Soley2025AttentiveGRUAECode}.

{\small
\bibliographystyle{unsrtnat}
\bibliography{ref}
}

\section{Supplementary Material}

\section*{S1.\quad Model Details and Notation}

\paragraph{Notation.}
Let $\{x_i\}_{i=1}^N$, $x_i\in\mathbb{R}^{T\times d}$ denote per-participant 28-day sequences of $d$ wearable features; $y_i\in\{0,1\}$ is the end-of-window depression label. The encoder outputs a sequence $\{h_t\}_{t=1}^{T}$ and a pooled embedding $h_i\in\mathbb{R}^{p}$.

\subsection*{S1.1\quad GRU encoder and attention}
For day $t$ with input $x_t$ and previous state $h_{t-1}$:
\begin{align*}
z_t &= \sigma(W_z[h_{t-1},x_t]),\qquad
r_t = \sigma(W_r[h_{t-1},x_t]),\\
\tilde h_t &= \tanh\!\big(W_h[r_t\odot h_{t-1},x_t]\big),\qquad
h_t = (1-z_t)\odot h_{t-1}+z_t\odot \tilde h_t 
\end{align*}

\noindent\textbf{Multi-head attention.} Two heads ($d_k{=}16$) are applied to $H=[h_1;\ldots;h_T]$.
Let $H\in\mathbb{R}^{T\times d_h}$. For head $m=1,\dots,M$,
\[
Q_m = H W_Q^{(m)},\quad K_m = H W_K^{(m)},\quad V_m = H W_V^{(m)},
\]
with $W_Q^{(m)},W_K^{(m)},W_V^{(m)} \in \mathbb{R}^{d_h\times d_k}$ (learned)
\[
A_m=\mathrm{softmax}\!\Big(\frac{Q_m K_m^\top}{\sqrt{d_k}}\Big)\,V_m
\]
Heads are concatenated and time-pooled to yield the participant embedding $h_i$.

\subsection*{S1.2\quad Decoder and reconstruction loss}
The decoder reconstructs $\hat X_i$ from $h_i$:
\[
\mathcal{L}_{\text{AE}}=\frac{1}{NTd}\sum_{i=1}^N\big\|X_i-\hat X_i\big\|^2
\]

\subsection*{S1.3\quad Outcome head and prediction loss}
A two-layer MLP on $h_i$ yields $\hat y_i=\sigma(f(h_i))$ with
\[
\mathcal{L}_{\text{BCE}}=-\frac{1}{N}\sum_{i=1}^N \Big[y_i\log\hat y_i+(1-y_i)\log(1-\hat y_i)\Big]
\]

\subsection*{S1.4\quad Joint objective and $(\alpha,\beta)$ selection}
\[
\boxed{\ \mathcal{L}=\alpha\,\mathcal{L}_{\text{AE}}+\beta\,\mathcal{L}_{\text{BCE}}\ }\,
\]
We performed a lightweight random search on the discovery split
($\alpha\in\{0.3,0.5,0.7,1.0\}$, $\beta\in\{0.5,0.7,1.0\}$; 5 draws),
selecting the setting that maximized validation AUC (Silhouette as tiebreaker).
Chosen: $(\alpha,\beta)=(0.7,1.0)$.

\subsection*{S1.5\quad Gradient “surgery” to reduce task conflict}
When $\langle g_{\text{AE}},g_{\text{BCE}}\rangle<0$, project away the conflicting component:
\[
g_{\text{AE}}\leftarrow g_{\text{AE}} - \frac{\langle g_{\text{AE}},g_{\text{BCE}}\rangle}{\|g_{\text{BCE}}\|^2}\, g_{\text{BCE}},\qquad
g \leftarrow g_{\text{AE}}+g_{\text{BCE}}.
\]
This stabilized joint training without altering the objective.

\subsection*{S1.6\quad Feature selection for model}
We restrict inputs to \texttt{sum\_duration\_asleep}, \texttt{avg\_duration\_asleep}, \texttt{avg\_efficiency},
\texttt{avg\_duration\_to\_fall\_asleep}, \texttt{first\_waketime}, and \texttt{first\_bedtime} to capture quantity,
quality, onset/offset (chronotype), and inertia of sleep over 28 days; sleep is a clinically established
correlate/modifier of depression; and limiting $d$ improves interpretability and reduces overfitting in
low-frequency data. Additional modalities are used downstream for subtype interpretation.

\FloatBarrier

\section*{S2.\quad Training, Clustering, and Validation Details}

\paragraph{Optimization.}
Adam (init LR $10^{-5}$, ReduceLROnPlateau), batch size 64, max 50 epochs, gradient clipping (norm 1.0),
dropout 0.3–0.5, $L_2{=}10^{-4}$; early stopping on validation AUC.

\paragraph{External validation.}
Freeze the encoder, embed DS3+DS4, reuse the trained GMM, and evaluate AUC/Silhouette. Cluster stability was assessed on external dataset using the ARI through a leave-one-out approach, randomly leaving out a small number of participants \( n \in \{1, \dots, 50\} \) from the external dataset and repeating the process 200 times to measure clustering robustness.

\paragraph{Selection for the number of clusters K.}
We fit GMMs for $K\in\{1,\ldots,9\}$ on discovery embeddings and select $K$ by BIC.

\begin{figure}[H]
  \centering
  \includegraphics[width=.55\linewidth]{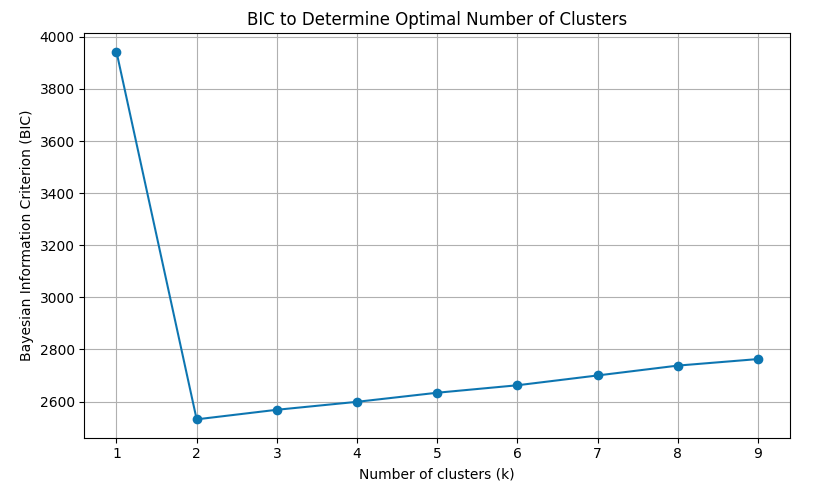}
  \caption{BIC across $K$. The elbow/minimum occurs at $K{=}2$. Models with $K{>}2$ did not improve AUC/Silhouette and produced small, unstable clusters.}
  \label{fig:bic}
\end{figure}
\FloatBarrier

\noindent \textit{Rationale.}
We retain $K{=}2$ for all ablations/baselines for comparability and because it aligns with prior outcome-guided temporal subtyping that often separates higher-risk vs.\ lower-risk disease trajectories \citep{Wang2024ClinicalTemporalClustering}, and with the two-group structure reported in GLOBEM-style depression analyses \citep{xu2023globem}. Specifically, peer‐reviewed studies have described “hyposomnic” (insomnia/short‐sleep) and “hypersomnic” (excessive/long‐sleep) depression subtypes \citep{Tsuno2005SleepDepression,Toenders2020NeurovegetativeSubtypes}. Data-driven studies in youth and adult cohorts likewise discover sleep-driven subtypes with neurovegetative contrasts \citep{Drysdale2017NeurophysSubtypes,Harvey2008BipolarSleepCircadian}. Our two clusters align with these well-established categories, strengthening biological plausibility and suggests some concrete first-line intervention (improve sleep regularity) rather than one-size-fits-all pharmacotherapy approach.

\section*{S3.\quad Additional Results and Plots}

\subsection*{S3.1\quad ARI stability under resampling}
\begin{figure}[H]
  \centering
  \includegraphics[width=.60\linewidth]{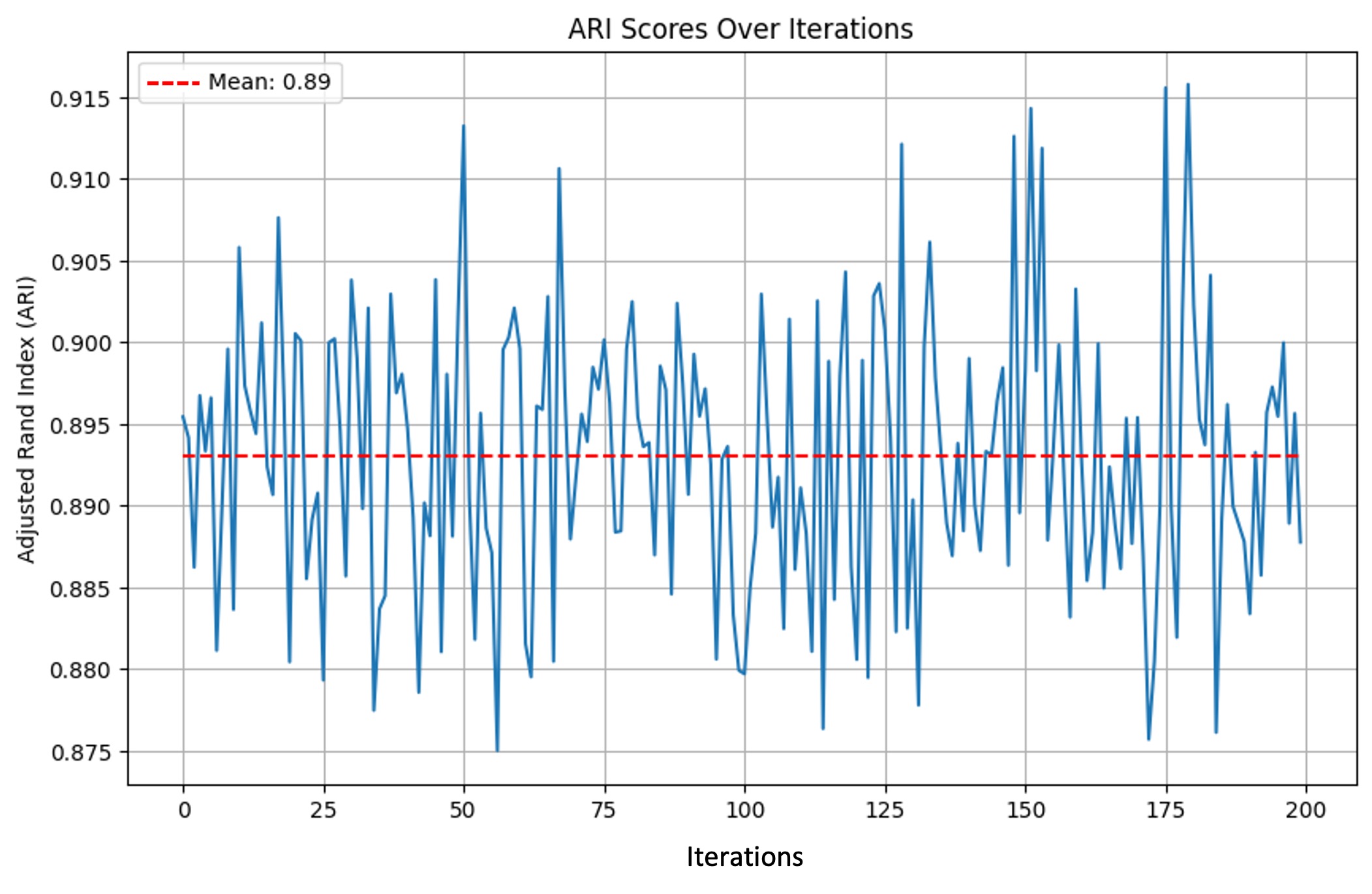}
  \caption{ARI over 200 resamples on DS3+DS4 (mean $\approx 0.89$, dashed line).}
  \label{fig:ari_sup}
\end{figure}
\FloatBarrier

\subsection*{S3.2\quad Top differentiating features (Cohen’s $d$)}
\begin{figure}[H]
  \centering
  \includegraphics[width=.65\linewidth]{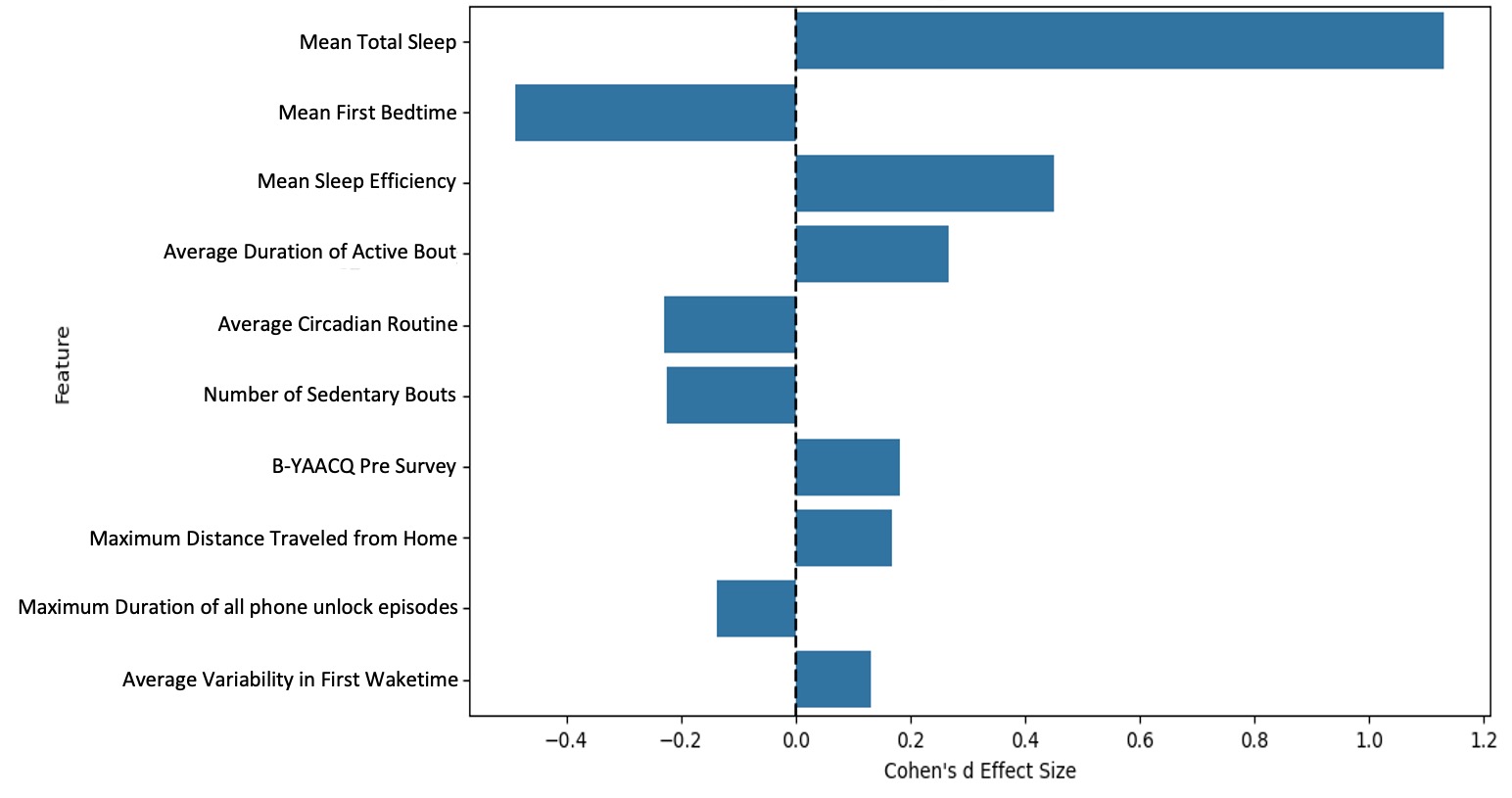}
  \caption{Top 10 features ranked by Cohen's $d$. Sleep features dominate.}
  \label{fig:cohens_bar_sup}
\end{figure}
\FloatBarrier

\subsection*{S3.3\quad Attention-based temporal interpretation}
\begin{figure}[H]
  \centering
  \includegraphics[width=.65\linewidth]{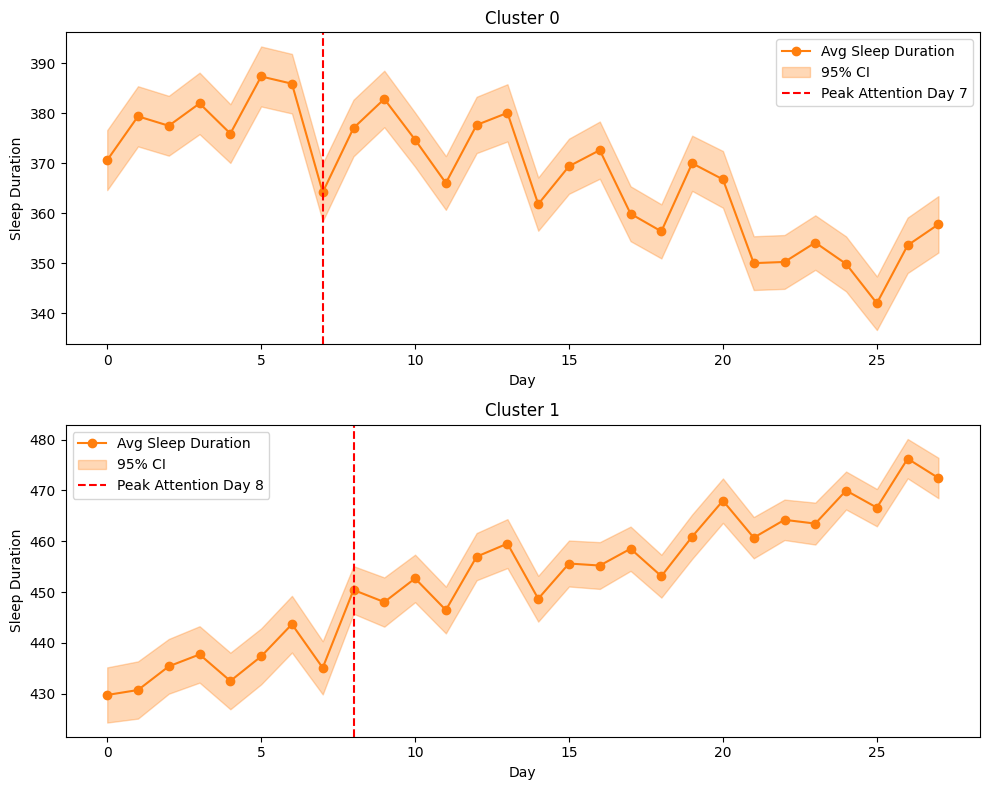}
  \caption{Cluster-level average sleep trajectories with peak attention day (red dashed).
  Cluster~0 peaks earlier (around day~7) near sharp declines; Cluster~1 peaks later (around day~8) as sleep stabilizes.}
  \label{fig:cluster_attention_sup}
\end{figure}

\begin{figure}[H]
  \centering
  \includegraphics[width=.65\linewidth]{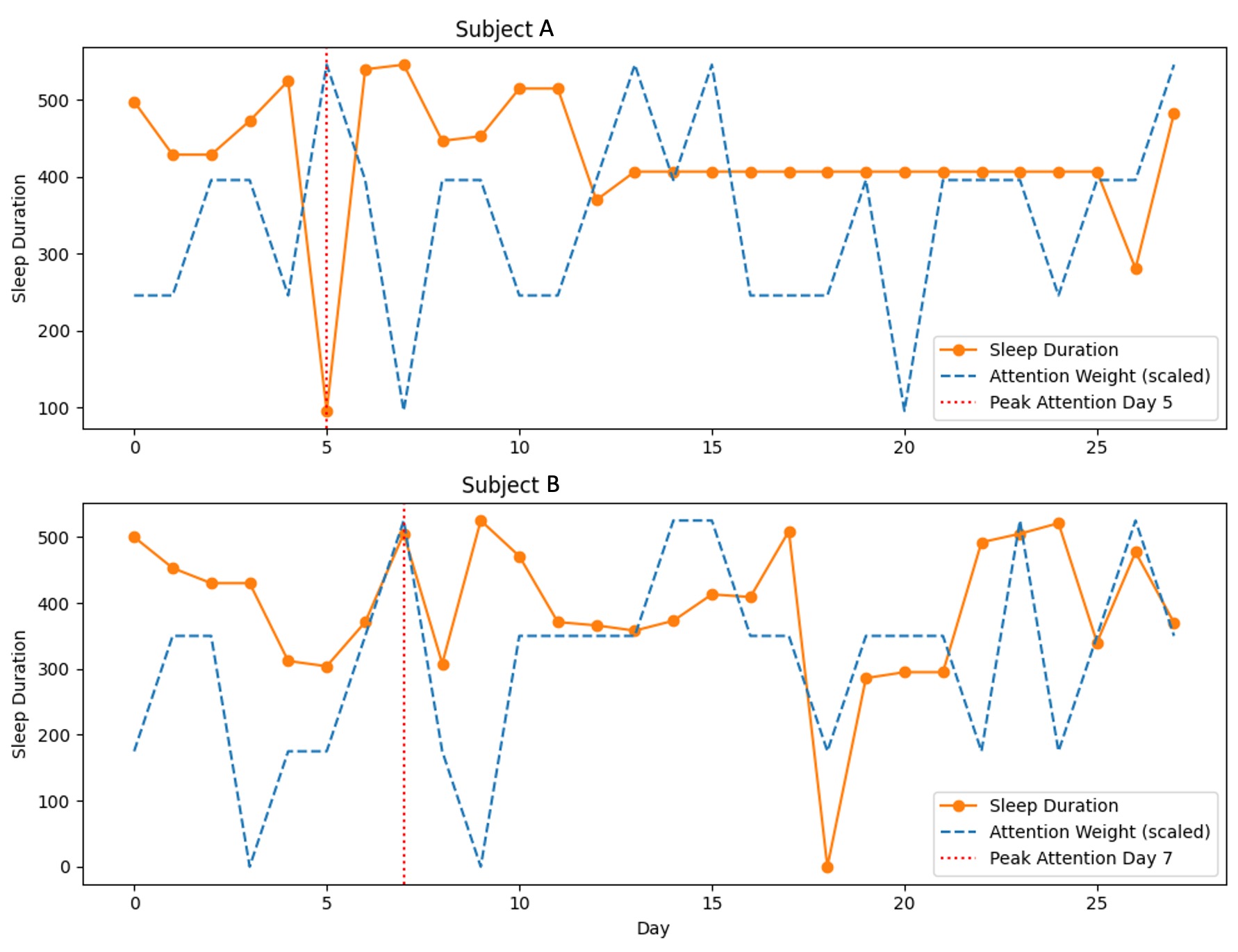}
  \caption{Attention weights over time (dashed blue line) for two participants A (cluster 0), B (cluster 1). Peak attention (vertical dashed red line) aligns with behavioral deviation (e.g., low, high sleep (solid orange line))}
  \label{fig:individual_attention_sup}
\end{figure}
\FloatBarrier

\subsection*{S3.4\quad Subtype summary (prevalence and peak attention)}
\begin{table}[H]
\centering
\caption{Subtype summary on discovery and validation: prevalence and average peak-attention day.}
\label{tab:subtype_summary}
\small
\setlength{\tabcolsep}{15pt}
\begin{tabular}{lcc}
\toprule
 & \textbf{Cluster 0} & \textbf{Cluster 1} \\
\midrule
Depressed fraction (disc.) & 55\% & 45\% \\
Peak attention (days)      & 5–7  & 8–10 \\
Qualitative profile        & lower/variable sleep & higher/stable sleep \\
\bottomrule
\end{tabular}
\end{table}
\FloatBarrier

\noindent Cluster~0 exhibits shorter total sleep, later bedtimes, and lower sleep efficiency; Cluster~1 shows longer, more stable sleep. Activity/survey signals are directionally consistent (longer active bouts, fewer sedentary bouts, and healthier BYAACQ patterns in Cluster~1). Attention peaks highlight inflection windows, which are earlier in Cluster~0 near declines; slightly later in Cluster~1 as stability returns.

\subsection*{S3.5.\quad Cluster-wise comparison of behavioral and survey features (Cohen’s $d$, FDR)}
We ran Mann–Whitney U tests per feature with Benjamini–Hochberg Adjusted-FDR correction and computed Cohen’s $d$.

\begin{table}[H]
\centering
\caption{Cluster-wise comparison of behavioral/survey features. Means $\pm$ SD. Effect sizes are Cohen’s $d$; $q$ is FDR- $p$ (top rows remain significant at $q{<}0.05$).}
\label{tab:all_features}
\scriptsize
\setlength{\tabcolsep}{5pt}
\begin{tabular}{lccc}
\toprule
\textbf{Feature} & \textbf{Cluster 0 (N=195)} & \textbf{Cluster 1 (N=177)} & \textbf{Cohen's $d$ \ / \ $q$} \\
\midrule
Average Total Sleep Duration & $371.65\pm77.09$ & $449.15\pm58.73$ & $1.13 \ / \ <0.001$\\
Average First Bedtime & $1535.89\pm125.75$ & $1487.09\pm64.62$ & $-0.49 \ / \ <0.001$\\
Average Sleep Efficiency & $92.85\pm4.89$ & $94.53\pm2.05$ & $0.45 \ / \ <0.001$\\
Average Active Bout Duration & $3.38\pm0.28$ & $3.48\pm0.42$ & $0.27 \ / \ <0.05$\\
Average Daily Circadian Routine & $0.53\pm0.11$ & $0.50\pm0.09$ & $-0.23 \ / \ <0.05$\\
Number of Sedentary Bouts & $57.06\pm7.86$ & $55.35\pm7.38$ & $-0.22 \ / \ <0.05$\\
Average Variability in First Bedtime & $91.55\pm78.73$ & $78.03\pm67.67$ & $-0.18 \ / \ <0.05$\\
BYAACQ (Pre) & $2.81\pm0.46$ & $2.92\pm0.72$ & $0.18 \ / \ <0.05$\\
Max Distance from Home & $42301.00\pm58967.92$ & $52913.61\pm66576.54$ & $0.17 \ / \ <0.05$\\
Average Variability in First Waketime & $90.86\pm77.76$ & $78.88\pm65.46$ & $-0.17 \ / \ <0.05$\\
Max Duration of Phone Unlocks & $53.32\pm19.14$ & $50.66\pm19.51$ & $-0.14 \ / \ <0.05$\\
PHQ-4 (weekly) & $3.12\pm0.15$ & $3.10\pm0.10$ & $-0.12 \ / \ <0.05$\\
BFI-10 (Openness, Pre) & $7.17\pm0.50$ & $7.05\pm0.43$ & $-0.12 \ / \ <0.05$\\
\midrule
PSS-10 (Pre) & $20.62\pm1.33$ & $19.76\pm1.88$ & $-0.09 \ / \ 0.227$\\
ERQ-Suppression (Pre) & $4.32\pm0.25$ & $4.28\pm0.25$ & $-0.14 \ / \ 0.323$\\
2-Way SSS (Receiving Emotional, Pre) & $25.00\pm1.15$ & $29.08\pm1.70$ & $0.05 \ / \ 0.193$\\
Average First Waketime & $518.63\pm152.74$ & $534.33\pm71.13$ & $0.13 \ / \ 0.976$\\
Time Spent at Home & $829.31\pm149.77$ & $813.28\pm125.53$ & $-0.12 \ / \ 0.129$\\
Average Sleep Latency & $0.15\pm0.90$ & $0.07\pm0.43$ & $-0.12 \ / \ 0.951$\\
STAI (Pre) & $45.87\pm2.11$ & $40.00\pm0.00$ & $-0.10 \ / \ 0.404$\\
CES-D-9 (Pre) & $8.37\pm0.86$ & $7.26\pm1.34$ & $-0.10 \ / \ 0.621$\\
SocialFit (Pre) & $74.17\pm2.30$ & $75.37\pm1.72$ & $0.10 \ / \ 0.387$\\
EDS (Pre) & $9.98\pm1.81$ & $10.16\pm2.17$ & $0.09 \ / \ 0.685$\\
Phone Unlock Episodes (count) & $102.22\pm38.21$ & $98.99\pm34.32$ & $-0.09 \ / \ 0.526$\\
Avg Time in Active Bout (min) & $186.75\pm29.64$ & $189.27\pm31.74$ & $0.08 \ / \ 0.842$\\
Total Sedentary Bout Duration (min) & $1250.74\pm31.75$ & $1253.25\pm29.64$ & $0.08 \ / \ 0.837$\\
Avg Sedentary Bout Duration (min) & $30.84\pm7.04$ & $31.28\pm7.29$ & $0.06 \ / \ 0.713$\\
CHIPS (Pre) & $20.04\pm4.50$ & $19.80\pm3.39$ & $-0.06 \ / \ 0.268$\\
BDI-II (Pre) & $14.34\pm1.74$ & $12.42\pm2.11$ & $-0.04 \ / \ 0.970$\\
Avg Step Count & $7380\pm10.27$ & $7500\pm21.31$ & $0.06 \ / \ 0.629$\\
FSPWB (Pre) & $43.09\pm1.32$ & $45.28\pm1.42$ & $0.14 \ / \ 0.494$\\
Avg Duration of Unlock Episodes (min) & $4.33\pm2.61$ & $4.29\pm3.61$ & $-0.01 \ / \ 0.156$\\
\bottomrule
\end{tabular}
\end{table}
\FloatBarrier

\end{document}